# Still no evidence for an effect of the proportion of non-native speakers on language complexity – A response to Kauhanen, Einhaus & Walkden (2023)

**Author:** Alexander Koplenig[1]

[1]Leibniz-Institute for the German Language (IDS), Mannheim, Germany.

E-mail: koplenig@ids-mannheim.de

**Abstract:** In a recent paper published in the Journal of Language Evolution, Kauhanen, Einhaus & Walkden (https://doi.org/10.1093/jole/lzad005, KEW) challenge the results presented in one of my papers (Koplenig, Royal Society Open Science, 6, 181274 (2019), https://doi.org/10.1098/rsos.181274), in which I tried to show through a series of statistical analyses that large numbers of L2 (second language) speakers do not seem to affect the (grammatical or statistical) complexity of a language. To this end, I focus on the way in which the Ethnologue assesses language status: a language is characterised as vehicular if, in addition to being used by L1 (first language) speakers, it should also have a significant number of L2 users. KEW criticise both the use of vehicularity as a (binary) indicator of whether a language has a significant number of L2 users and the idea of imputing a zero proportion of L2 speakers to non-vehicular languages whenever a direct estimate of that proportion is unavailable. While I recognise the importance of post-publication commentary on published research, I show in this rejoinder that both points of criticism are explicitly mentioned and analysed in my paper. In addition, I also comment on other points raised by KEW and demonstrate that both alternative analyses offered by KEW do not stand up to closer scrutiny.

# 1. Introduction

The linguistic niche hypothesis proposes that the social niche a language occupies in a community affects its structural properties. Specifically, according to the linguistic niche hypothesis, languages with large numbers of speakers tend to simplify their grammar and have a reduced structural complexity (Lupyan & Dale 2010; Dale & Lupyan 2012).

The linguistic niche hypothesis assumes that languages that are spoken by more people over greater geographic areas will, on average, also be learned by a greater proportion of L2 learners, i.e. often adults. Since complex morphology appears to be difficult to learn for adults, the linguistic niche hypothesis conjectures that there should be a negative selection over time against such hard-to-learn morphological paradigms for languages with a larger number of L2 speakers compared to languages that are mainly learned during childhood as L1. This, in turn, it is argued, explains the observed negative statistical association between speaker population size and morphological complexity (Dale & Lupyan 2012; Trudgill 2001). In a recent study (Koplenig 2019a), I have pointed out that since the argument outlined above is inductive by nature, its validity cannot be simply taken (more or less implicitly) for granted. Crucially, Lupyan & Dale (2010) use the estimated speaker population size as a proxy for the proportion of L2 speakers (Nettle 2012). In my paper, I tested this conjecture empirically for more than 2,000 languages and showed that the results question the idea of the impact of non-native speakers on the grammatical and statistical structure of languages.

The main obstacle in this context is the fact that, as Bentz & Winter (2013) point out, estimations regarding a breakdown of L1 versus L2 populations are very limited. In general, most information regarding speaker population sizes/compositions are based on the Ethnologue (Simons & Fennig 2017). The Ethnologue categorizes each language in regard to how endangered it is using the Expanded Graded Intergenerational Disruption Scale (EGIDS).

In this context, a language is categorized as vehicular if it is used as an L2 in addition to being used as an L1. This information can be used to indirectly gain information about the proportion of L2 users: "A language at EGIDS 4 or lower is, by definition, a local language and L2 users are not expected. However, languages at EGIDS 3 and higher are vehicular and, by definition, they should have a significant number of L2 users" (Simons & Fennig 2017). The great advantage here is that information of the EGIDS level is available for all languages that are listed in the Ethnologue. In my paper, I used vehicularity as an indicator for whether a language is used by L2-speakers, to test the assumed relationship between the L2 proportion and both morphological and information-theoretic complexity. Through a series of statistical analyses, I tried to show that large L2 proportions do not seem to affect the (grammatical or statistical) complexity of a language.

In a recent comment published in the Journal of Language Evolution, Kauhanen, Einhaus & Walkden (2023; KEW) challenge my findings. KEW criticise both the use of vehicularity as a (binary) indicator of whether a language has a significant number of L2 users and the idea of imputing a zero proportion of L2 speakers to non-vehicular languages whenever a direct estimate of that proportion is unavailable. While I recognise the importance of post-publication commentary on published research, I will show in what follows that that both points of criticism are explicitly mentioned and analysed in my paper. In addition, I will also comment on other points raised by KEW and demonstrate that both alternative analyses offered by KEW do not stand up to closer scrutiny.

## 2. Using vehicularity to test the linguistic niche hypothesis

KEW criticise the use of vehicularity as a proxy of whether a language is likely to have significant numbers of L2 speakers. They state (p. 3): "In Koplenig's analysis, languages with an EGIDS score of 3 or lower are defined to be vehicular, the rest being non-vehicular". I believe it is important to point out that this not my definition or mapping, but the way how languages are categorized by the Ethnologue. KEW (p. 3) rightfully point out that a "considerable number of non-vehicular languages are reported by Ethnologue to be used as an L2 even though no numerical estimate of L2 users is given." I fully agree that this inconsistency is problematic and that it is thus important to ask if vehicularity is a good proxy for whether a language is used as an L2. I explicitly discuss this in the concluding section of my paper. Here, I quote the editors of the Ethnologue: "Based on the use of the phrase "vehicular language" by some as a synonym for lingua franca, we use the term vehicular to refer to the extent to which a language is used to facilitate communication among those who speak different first languages. If a language is characterized here as being Vehicular, it is used by others as an L2 in addition to being used by the community of L1 speakers." (Lewis & Simons 2010; see also Figure 1 therein). Based on this assessment, I believe that it is appropriate to use vehicularity in order to test the linguistic niche hypothesis: a language that is defined as vehicular should – according to the Ethnologue – be a language that is "used for communication between strangers" (KEW, p. 1; also see Wray & Grace 2007) and that "should have a significant number of L2 users" (Simons & Fennig 2017, p. 20). Thus, if the linguistic niche hypothesis holds, we should expect that there is a statistical association between vehicularity and complexity: In my paper, I show that this is not the case for either morphological or information-theoretic complexity, when controlling for estimated speaker population size.

Importantly, the problem that there are non-vehicular languages for which Ethnologue reports a proportion of L2 users greater than 0 is not concealed by me, but explicitly mentioned in section 2.2 of my paper and – as also mentioned in the paper – additional analyses are presented and discussed in section 7 of the electronic supplementary material, where languages categorised as non-vehicular but with L2 proportions greater than zero are removed from the analyses. The reported results generally support the results presented in the main part of the paper.

**3. Using vehicularity to impute L2 proportions**

In a set of further statistical analyses, I used vehicularity to impute missing values: In correspondence with the categorization scheme of the Ethnologue (Simons & Fennig 2017), non-vehicular languages with no available information on L2 users are assigned an L2 proportion of 0. KEW are right to point out that this step is worth discussing since this zero-imputation strategy affects almost all non-vehicular languages. Importantly, however, imputed values are only used for the non-parametric Spearman correlation analyses. Here, I test whether there is a significant (determined by non-parametric permutation tests) monotonic relationship between (morphological or information-theoretic) complexity and the L2 proportion after removing the effect of speaker population size and vice versa (correlating complexity and speaker population size while controlling for the L2 proportion). Since – as mentioned in the paper, section 2.6 – Spearman correlation coefficients and part Spearman correlation coefficients can be computed as Pearson's correlation coefficient on the ranks of the two variables, the zero-imputation strategy implies that all non-vehicular languages are assigned the lowest rank in each analysis, an assumption that I believe is reasonable, but worthy of discussion. KEW (section 5.1) present three so called complete case analyses, where all cases with missing information are removed (no imputation). In parametric linear mixed effects (or multilevel) models, (morphological or information-theoretic) complexity is predicted by fixed effects of the (log) population size and the L2 proportion and a random intercept for language family. Additionally, the two models with morphological complexity as the outcome also include a random intercept for linguistic area. Based on their results (Table 2 – Table 3), KEW argue that "population size and the proportion of L2 speakers have a declining effect on morphological complexity, and both predictors are statistically significant." (p. 6). For information theoretic complexity as the outcome (Table 4), KEW find no evidence "for an effect of either the proportion of L2 speakers or population size" (p. 6). There are two major methodological problems with KEW's models: (i) KEW do not include any random slopes in their models, due to convergence issues with the R package they are using to fit linear mixed effects models. (ii) Estimates in KEW's models are derived by Restricted Maximum Likelihood (REML). This is highly problematic, because they use Akaike's information criterion (Akaike 1974, AIC) for model selection. For example, on p. 6, KEW argue that they "do not include interactions between the covariates in any of our models, as doing so always leads to a worse model when quantified on AIC". However, when different sets of fixed effects are considered, estimates must not be derived by REML, but by Maximum Likelihood (ML) (Verbeke & Molenberghs 2001; Zuur et al. 2009; Faraway 2016). To solve both (i) and (ii), I generated a set of 72 candidate models consisting of all possible combinations of fixed effects for speaker population size, the L2 proportion and their interaction, crossed random intercepts for language family and linguistic area and random slopes for both speaker population size and L2 proportion for both random effects. As outcome, I consider information theoretic complexity and two versions of morphological complexity: (i) A full version that incorporates all languages that have information for at least one available of a total of 28 features from the Word Atlas of Language Structures (WALS) (Dryer & Haspelmath 2013) that are used to construct the index of morphological complexity, (ii) a version for a subset of languages with available information for at least six features (for details regarding the index construction, see my paper or KEW, section 2). I use Stata/MP 17 for the linear mixed effects models, estimates were derived by ML and models were fitted with gradient-based maximization. Of a total of 216 models, 212 or 98.15% converged to an optimal solution, thus pointing towards problems in KEW's analyses. Table 1 summarizes the results. I first checked for all three outcome versions whether the model with the lowest AIC includes the L2 proportion as a fixed effect, row 4 of Table 1 shows that this is only the case for the versions where morphological complexity is the outcome. This means that including the speaker L2 proportion does not improve the model fit in case of information-theoretic complexity. I then extracted the best models that include a fixed effect for the L2 proportion. Rows 4 – 6 list the model structure per outcome. Row 7 shows that there is only evidence for a significant effect of the L2 proportion for morphological complexity as the outcome, when only languages with available information for at least six WALS features are considered, for both other outcome versions, there is no evidence for a significant effect of the L2 proportion on complexity at any standard level of significance.[1]

However, as pointed out in my paper and discussed further in the accompanying supplementary material for a similar sample, is not clear whether any of the three samples are unbiased: (i) row 8 of Table 1 shows that almost all languages in all three samples have an L2 proportion that is greater than zero with a median estimate (row 9) of more than 15%, which seems rather high given the assumption of the linguistic niche hypothesis that most languages have almost no L2 speakers.

---

[1] The results obtained are not altered when an alternative version of Aikaike's information criterion, AICc (Hurvich & Tsai 1989), is used instead of AIC. AICc accounts for sample size by including an additional bias correction term, see also Burnham & Anderson (2004).

| Row | | Outcome | | |
|---|---|---|---|---|
| | | MC | | H |
| 1 | Minimum number of included WALS features/chapters | 1 | 6 | - |
| 2 | Does the best model include a fixed effect for the L2 proportion | yes | yes | no |
| 3 | Fixed effects | Population, L2 proportion and their interaction | Population and L2 proportion | Population and L2 proportion |
| 4 | Random intercepts | Family and Area | Family and Area | Area |
| 5 | Random slopes | Population and L2 proportion for Family | L2 proportion for Family and Area | - |
| 6 | Estimated effects for the best model that includes a fixed effect for the L2 proportion, shown are the estimated beta coefficients for each fixed effect (parametric $p$-values in parentheses) | $\beta_{logPop} = -0.010$ ($p = 0.141$) $\beta_{L2prop} = -0.067$ ($p = 0.732$) $\beta_{interaction} = -0.026$ ($p = 0.141$) | $\beta_{logPop} = -0.013$ ($p = 0.006$) $\beta_{L2prop} = -0.217$ ($p = 0.001$) | $\beta_{logPop} = 0.037$ ($p = 0.002$) $\beta_{L2prop} = -0.108$ ($p = 0.338$) |
| 7 | Number of languages in the sample | 144 | 97 | 80 |
| 8 | How many languages in the sample have an L2 proportion > 0 | 97.22% | 95.88% | 100.00% |
| 9 | Median L2 proportion | 0.16 | 0.16 | 0.18 |
| 10 | Median speaker population size | 786,500 | 1,480,000 | 4,611,750 |
| 11 | Spearman correlation between speaker population size and the L2 proportion (non-parametric permutation $p$-value in parentheses) | -0.082 ($p = 0.276$) | -0.030 ($p = 0.748$) | -0.105 ($p = 0.360$) |

**Table 1 – Overview of the results of the complete case linear mixed effects model analyses for each outcome. $\beta_{logPop}$ – estimated coefficient for the log of speaker population size; $\beta_{L2prop}$ – estimated coefficient for the L2 proportion; $\beta_{interaction}$ – estimated coefficient for the interaction between the L2 proportion and the log of speaker population size.**

(ii) Compared to the median estimated speaker population size of 7,000 for the all languages listed by the Ethnologue (Lupyan & Dale 2010), the median estimated speaker population size in the samples ranges between 786,500 and 4,611,750 (row 10). To put this into perspective, I drew 1,000,000 random samples of all 6,880 living languages listed in the Ethnologue (Simons & Fennig 2017) with each sample consisting of 100 languages. For each sample, I computed the median population size. The median estimate for the median population size is 8,000 and the sample with the highest median population size was 51,000. Thus, the three samples used for KEW's complete cases analyses have a median population size that is several orders of magnitude larger than what we would expect from a random sample, and the probability of randomly drawing a sample like the one used in the complete cases analyses above is less than one in a million. This calls into question the appropriateness of standard parametric frequentists approaches (Koplenig 2019b; Koplenig 2021) which is why in my original paper, I used non-parametric tests that do not make any assumptions regarding the stochastic mechanism that generated the data (Freedman & Lane 1983b). (iii) There is no positive Spearman correlation between the estimated size of the speaker population and the estimated proportion of L2 speakers in any of the three samples (row 11). As I have written above and in my paper, this is actually a key assumption of the linguistic niche hypothesis.[2]

As an alternative to frequentist analyses, I now adopt a Bayesian perspective and fit Bayesian multilevel models with fairly uninformative priors[3] for each "best" model that includes a fixed effect for the L2 proportion (see Table 1, Rows 4 – 6). Figure 1 visualizes the results. In all but one cases, the 95% equal tailed credible interval contains zero. This suggests that there is not enough evidence to support the hypothesis that these coefficients are different from zero.

---

[2] In two recent papers (Koplenig 2021; Koplenig, Wolfer & Meyer 2023, Supplementary Information S12), I have explored the problem of small samples in more detail. I show that such samples tend to be biased towards languages with more speakers. I argue that this has important implications for any quantitative typological investigation, since it can be shown that analyses of smaller corpora can lead to potentially incorrect conclusions.

[3] In Bayesian analysis, all coefficients are considered random variables, including those that are commonly called "fixed." These coefficients are assigned normal priors with zero mean and a variance of 10,000. Random intercepts, random coefficients, and error variances are assigned inverse-gamma priors with a shape of 0.01 and a scale of 0.01. For each model, I simulated three chains. I used a burn-in period of 10,000 for information-theoretic complexity as the outcome, and a longer burn-in period of 200,000 for the two models with morphological complexity as the outcome due to the more complex model structure.

The exception is the model with information-theoretic complexity as the outcome, here there is evidence for a significant positive effect of speaker population size on complexity. This is neither true for the proportion of L2 speakers in any of the three models, nor for the interaction between speaker population size and the proportion of L2 speakers for the full model with morphological complexity as the outcome. From a Bayesian perspective, we thus find no support for the linguistic niche hypothesis.

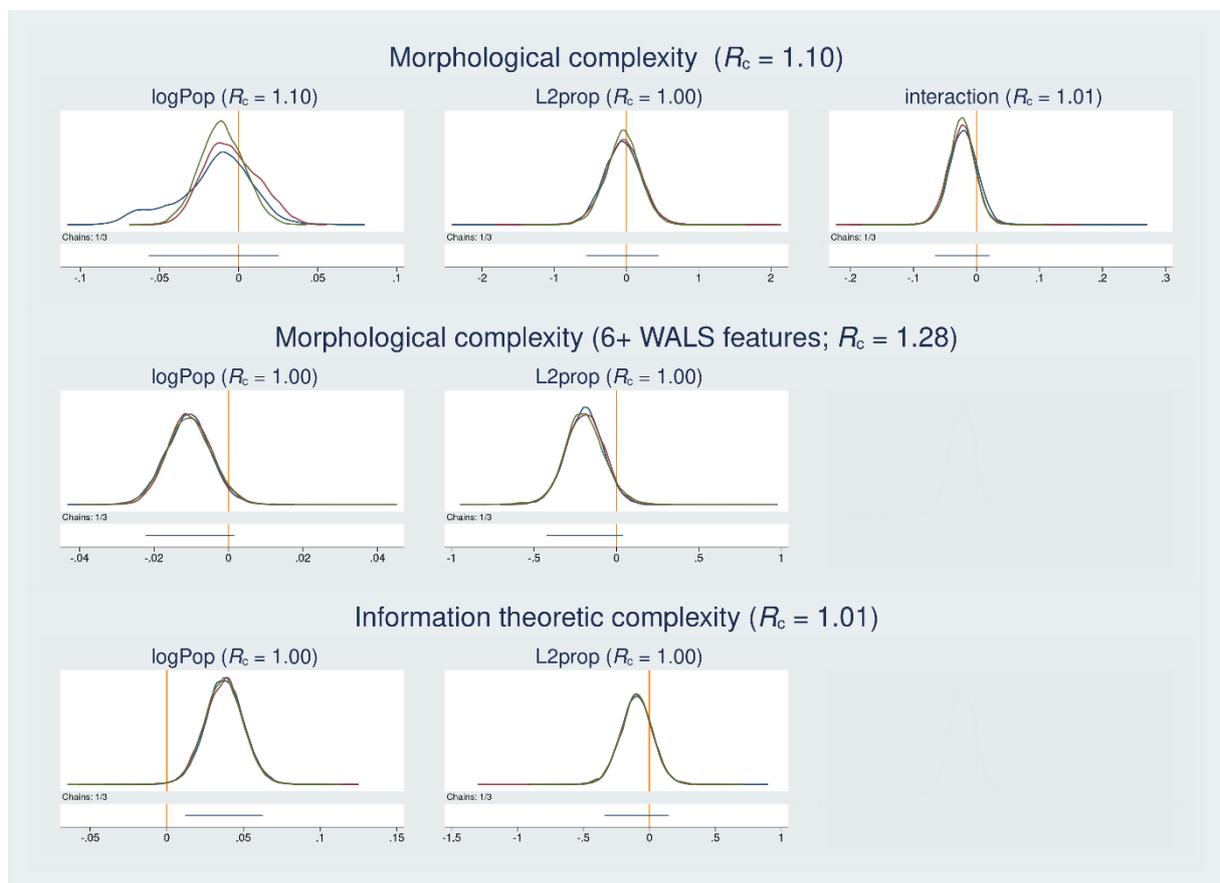

**Figure 1: Results of the Bayesian multilevel models, each row visualizes the results for one of the three outcome specifications, while each column plots the kernel density of the simulated marginal posterior distributions for each estimated parameter. In each case, the small plots on the bottom visualize the 95% equal-tailed credible interval. The Gelman–Rubin convergence diagnostic (Gelman & Rubin 1992; Brooks & Gelman 1998), $R_c$, is given in parentheses for each model and for each estimated parameter. While most values are below 1.2 indicating convergence, the overall $R_c$-value for the model with morphological complexity as the outcome for the subset of languages with available information for at least six WALS features is a bit high. However, for all three models, each $R_c$ value is below 1.2 for all estimated fixed effects.**

Against this background, I would like to reiterate that KEW are right to point out that the imputation step is worth discussing, I do so myself in section 2.2 of my paper and present complete cases analyses without any imputation (similar to that presented by KEW, but without making parametric assumptions) in section 10 of the supplementary material.

## 3. Using multiple imputation to impute L2 proportions

In their second set of analyses, KEW utilized a technique called multiple imputation to fill in the missing data (Rubin 1976; Rubin 1987; van Buuren 2018). While the idea of imputing data in a scenario with a missingness rate of ~92% (KEW, p. 10) may seem very desirable at first glance, upon closer examination, it becomes clear that KEW's approach suffers from severe and systematic biases that render it an unreliable method for accurately filling in missing data in the present scenario. Nonetheless, I am grateful to KEW for introducing me to this rather interesting technique. Multiple imputation offers a flexible, simulation-based technique to handle missing data consisting of three steps (StataCorp 2021): (i) setting up an imputation model and generating $m$ imputations (completed datasets), in KEW's analyses, $m = 100$; (ii) the completed datasets are separately analysed with standard statistical techniques; (iii) the results obtained in (ii) are pooled to provide estimates of the parameters of interest, accounting for the uncertainty due to missing data. KEW rightfully point out that the technique is based on the assumption that the data are either missing completely at random (MCAR), i.e. missingness for a variable $x$ is unrelated to the observed values of both other variables and the unobserved values of $x$, or missing at random (MAR), i.e. missingness on $x$ is uncorrelated with the unobserved value of $x$, after other variables in the dataset have been used to predict missingness on $x$. Or put differently: after controlling for the observed variables, the probability of missingness is independent of the true value of $x$. If this is not true, the data are said to be missing not at random (MNAR), if the value of $x$ itself predicts missingness (Medeiros 2016; UCLA: Statistical Consulting Group 2023). In general, multiple imputation methods assume that data are MCAR or MAR and not MNAR. KEW (p. 11) argue that "In the current state of understanding, we feel it would be premature to conclude one way or the other, but we point out that any argument to the effect that L2 speaker proportions are MNAR would need to specify a mechanism whereby such missingness arises." Unfortunately, there is no formal test to answer this question, since, as written above, the data that would be needed to determine this are themselves missing. However, there is a different way of representing MAR (Bartlett 2012): MAR implies that the distribution of $x$, given our imputation model, is the same whether or not $x$ is observed. If the data are MNAR, however, the chance of observing a value of $x$ depends on $x$, even after conditioning on our model. In that case, given our imputation model, the observed data does not "tell us how the missing values differ from the observed values" (Bartlett 2012). As an imputation model ($M_{KEW}$) to impute L2 proportions, KEW use a mixed effects model with the (logit-transformed) L2 proportion as outcome, random intercepts for (either morphological or information-theoretic) complexity, the log of population size and the log of range size. Now, the question is whether the chance of observing a value of the L2 proportion really does not depend on this value itself after conditioning on $M_{KEW}$. I would argue that it does, or do we really believe that $M_{KEW}$ tells us *all* there is to know regarding the question how the missing values of the L2 proportions are different from the corresponding observed values? KEW (p. 11) themselves sketch one such mechanism: "greater proportion of L2 speakers in a speech community increases, in general, the access that outsiders have to that community, and hence also increases the likelihood of the demographic variable of L2 speaker proportion being recorded by field typologists." This seems like a reasonable assumption: the smaller the value of the L2 proportion, the bigger the chance that this value is missing. I would even say that missing L2 proportion could be used as standard example of an MNAR type, since we almost exclusively observe higher values of L2, as written by KEW (p. 3); only in four cases, Ethnologue provides "an actual numerical zero proportion estimate." But, since there is no test to formally determine this, different researchers can have different opinions regarding the type of missingness. Nevertheless, we can test the efficiency of $M_{KEW}$ in order to find out if the multiple imputation will produce unbiased estimates, even in the presence of large proportions of missing data. KEW (p. 11) cite Madley-Dowd et al. (2019) who demonstrate that missingness up to "90% is tolerated by the method as long as the imputation model includes all necessary predictors". Madley-Dowd et al. (2019) convincingly show that this strongly depends on the strength of the imputation model. In their simulation study, strength is determined using the coefficient of determination $R^2$ that measures the proportion of variance in the outcome that is predictable by the imputation model. To provide unbiased estimates in the presence of high rates of missingness, Madley-Dowd et al. (2019) show that the imputation model needs to be almost perfect with an $R^2$ as high as 92%. To test this for $M_{KEW}$, I re-ran the specified models and computed an $R^2$ of ~44% for morphological complexity as a covariate in $M_{KEW}$ and corresponding $R^2$ of ~32% for information-theoretic complexity as a covariate in $M_{KEW}$.[4] If we compare this with the results of Madley-Dowd et al. (2019, Table 2), we find out that with a missingness rate of 90%, the reduction in standard error (compared to a complete cases analysis model) is arguably very unimpressive, ranging between ~0% and ~9%.[5] As a guide to test efficiency gains, Madley-Dowd et al. (2019) show that the fraction of missing information (FMI) is a valuable quantity (ranging between 0 and 1) to determine potential efficiency gains from multiple imputation: "The FMI is a parameter-specific measure that is able to quantify the loss of information due to missingness, while accounting for the amount of information retained by other variables within a data set" (Madley-Dowd et al. 2019, p. 64). Its interpretation is similar to an $R^2$, so an FMI of, say 0.2 means that 20% of the total sampling variance can be attributed to missing data. A high value indicates a problematic variable (UCLA: Statistical Consulting Group 2023). KEW report FMI values for both their imputations models, but do not interpret them: for information-theoretic complexity as covariate (Table 6), the FMI value is ~62% and for morphological complexity as covariate (Table 5), the FMI value is ~89%. This alone shows that $M_{KEW}$ does not provide much information about the missing values, especially in case of morphological complexity, where KEW report a negative significant effect of the L2 proportion on complexity. For information-theoretic complexity, there is no indication of an effect at any standard level of statistical significance. To further investigate this, I extracted all 100 imputed completed samples based on $M_{KEW}$ from the code provided by KEW. First, I computed the Spearman correlation between the imputed L2 proportion and speaker population size for each completed dataset and for both types of complexity.

---

[4] For such a mixed effects model, I computed the so-called conditional $R^2$, i.e. the variance explained by both the fixed effects and the random intercept, as suggested by Nakagawa & Schielzeth (2013). Note that in the fully observed data, there is information for morphological complexity as a covariate in the imputation model for only 28 language families. However, as KEW write (p. 8), their model imputes information for 122 families. This means that ~77% of all language families are systematically missing. Similar quantities are obtained for information-theoretic complexity, where information is systematically missing for ~84% of all language families. To put this into perspective, I computed a linear mixed effects model with the log of population size as the outcome and a random intercept for family for all available data points ($N = 2,143$). The variance of the random intercept is 6.63. I then re-ran the model, but restricted the computation to data points where the L2 proportion is non-missing ($N = 171$). Here, the variance of the random intercept is about 3.5 times higher, with a value of 24.10. Jolani (2018), who developed the imputation method used by KEW, discusses potential biases in the estimated random effects parameters that can arise from systematically missingness. He presents simulation results for systematically missing rates of up to 30%; it does not seem unlikely that problems could be more pronounced for missingness rates of more than 75%.
[5] To be exact, for $R^2 = 52$% the error reduction is 8.86%, for $R^2 = 40$% the error reduction is 2.18% and for $R^2 = 36$% the error reduction is 0.11%.

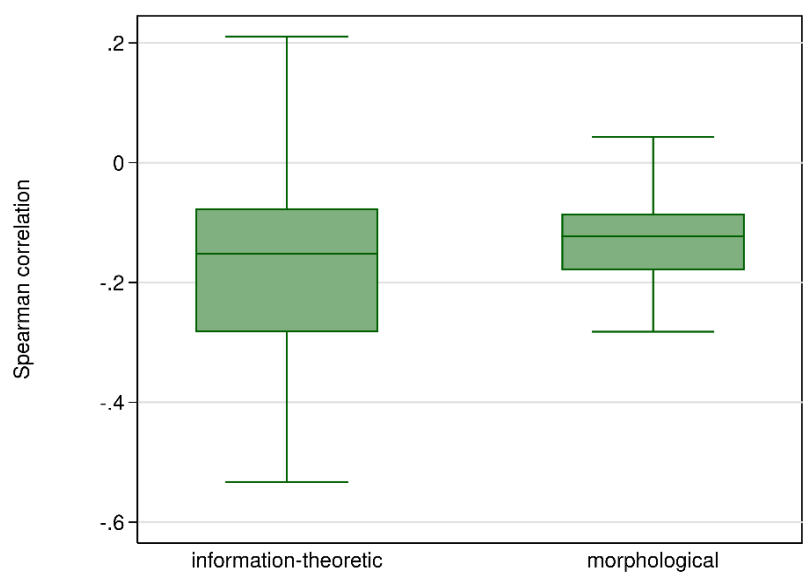

**Figure 2: Spearman correlation between the imputed L2 proportion and speaker population size per type of complexity (information-theoretic or morphological) each across 100 completed samples from KEW's multiple imputation analysis.**

Figure 2 presents the results: for both types of complexity, the Spearman correlation is negative in most samples, in 75% of all samples the Spearman correlation is lower than -0.08/-0.09 for information-theoretic/morphological complexity. This seems rather implausible and – as written above and in my paper – contradicts a basic assumption of the linguistic niche hypothesis.

For each sample, I then computed the percentage of languages that have an L2 proportion of (i) more than 0, (ii) more than 0.10, (iii) more than 0.25 and (iv) more than 0.50. Figure 3 visualizes the results. For information-theoretic complexity, all languages have an L2 proportion > 0, for morphological complexity, the median across samples is 99,74%. Both results do not seem plausible with respect to the linguistic niche hypothesis. For the remaining quantities, the results seem to be equally implausible: for information-theoretic complexity, the median percentages are 69.89% for the L2 proportion > 0.10, 49.44% for the L2 proportion > 0.25 and 30.74% for the L2 proportion > 0.50; for morphological complexity, the median percentages are 59.48% for the L2 proportion > 0.10, 44.47% for the L2 proportion > 0.25 and 31.57% for the L2 proportion > 0.50. In my view, it is very hard to argue that a model that assumes that almost a third of all languages have an L2 proportion of over 50% reflects the linguistic reality.

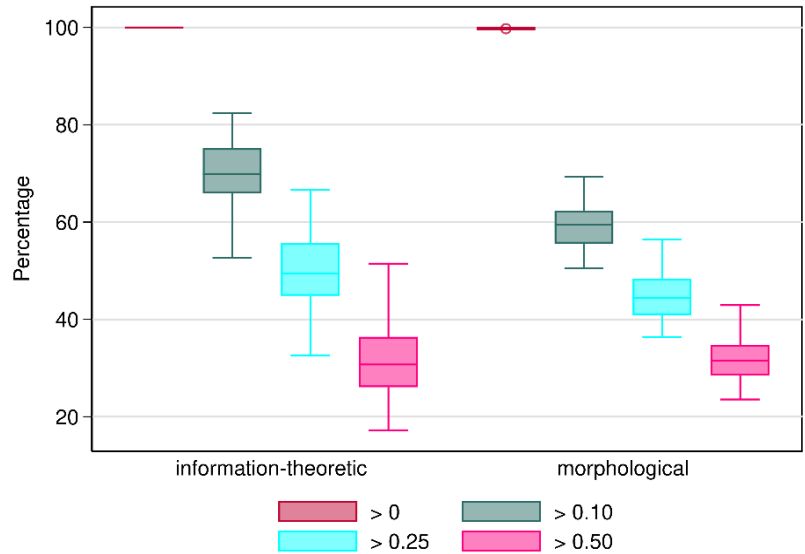

**Figure 3: Percentage of languages that have an L2 proportion of (i) more than 0, (ii) more than 0.10, (iii) more than 0.25 and (iv) more than 0.50 per type of complexity (information-theoretic or morphological) each across 100 completed samples from KEW's multiple imputation analysis.**

Finally, I computed the median estimated L2 proportion in each sample for both non-vehicular and vehicular languages. Figure 4 presents the results: for both types of complexity, completed datasets based on $M_{KEW}$ assume a lower L2 proportion for vehicular languages (median for information theoretic complexity = 0.20, for morphological complexity = 0.17) than for non-vehicular languages (median for information theoretic complexity = 0.26, for morphological complexity = 0.16). As written above and in my paper, this completely contradicts the categorisation scheme of the Ethnologue and basic typological intuitions.

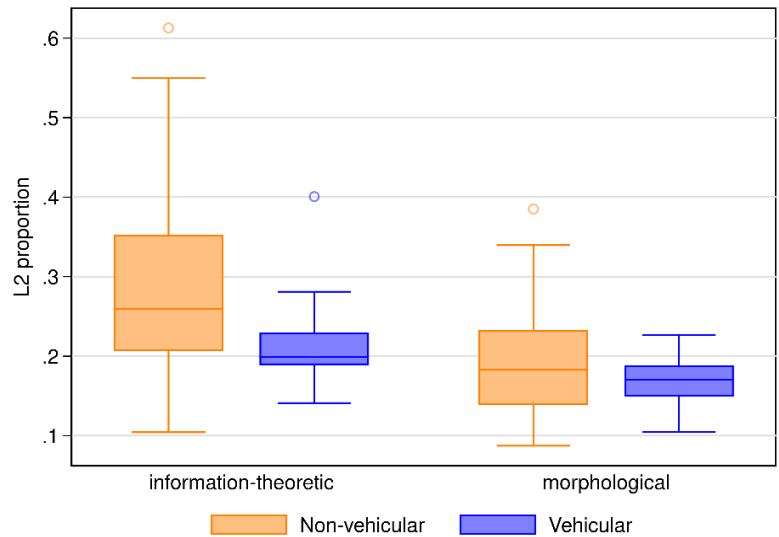

**Figure 4: Median L2 proportion for non-vehicular and vehicular languages per type of complexity (information-theoretic or morphological) each across 100 completed samples from KEW's multiple imputation analysis.**

## 4. Concluding remarks

In sum, I would like to thank KEW for giving me the opportunity to revisit the relationship between language complexity and the proportion of non-native speakers. However, as pointed out above, I have already addressed their two main points of critique in my original paper. In this context, I would like to thank one of the reviewers of my original paper, to whom I owe the consideration of these two points of criticism - she pointed out that both "the relationship between the two variables and the fact that not all languages with a vehicularity of 0 have 0 L2 speakers needs to be dealt with openly" and that there needs to be an analysis that "removes the 78 languages with a vehicularity index of 0 and a proportion of L2-speakers > 0". Here I refer the interested reader to the review reports, which I have deliberately chosen to make freely available online.

Against this background, I hope to have demonstrated convincingly that the two alternative analyses offered by KEW do not stand up to closer scrutiny: (i) Only one in three linear mixed effects model analyses based on complete cases supports the linguistic niche hypothesis at all, and there are good reasons to doubt the appropriateness of the samples used to test for an effect of L2 proportions on complexity. In addition, evidence for a significant effect of the L2 proportion is only found in standard-frequentist setting. Additional Bayesian multilevel models do not support this result. (ii) The multiple imputation analyses suffer from similar biases, and it is clear from the interpretation of the FMI values reported by KEW that the imputation model does not provide much information about the missing values.

Nevertheless, KEW and I might agree that neither non-imputation nor imputation of L2 proportions is an ideal strategy. This brings us back to the use of vehicularity as an indicator of high L2 vs. low L2 languages. To drive home my point, let me give an illustrative example that shows why it is possible to use such proxy variables to test claims between continuous variables. In a study examining the relationship between occupational exposure to a certain chemical and the risk of developing a specific health outcome, researchers may want to accurately measure the level of exposure to the chemical for each participant. However, it may be challenging to obtain accurate measurements of exposure, especially if exposure occurred in the past or if the exposure was intermittent. Instead, they may rely on the assessment of medical experts, such as occupational health physicians, to classify each participant as having either high or low exposure based on their job history, work practices, and other relevant factors. While this strategy does not seem to be perfect and information certainly gets lost when turning a continuous variable into a categorical one, it seems justified to statistically compare the incidence of the specific health outcome between the high and low exposure groups to assess whether there is a significant association between exposure to the chemical and the risk of developing the health outcome.

In our scenario, the exposure to the chemical element is the proportion of non-native speakers, the outcome is language complexity and the occupational health physicians are the field linguists of the Ethnologue that classify languages into high L2, i.e. vehicular languages and low L2 language, i.e. non-vehicular languages. Given that the sample of languages for which we have available information cannot be considered a random sample of the population of all existing languages as I have argued above and elsewhere (Koplenig 2019b; Koplenig 2021), I have used a non-parametric permutation testing approach that does not make any parametric assumptions about the data generating process (Freedman & Lane 1983a; Freedman & Lane 1983b; Berk & Freedman 2003). In section 5 of the supplementary information of my paper, the permutation test is validated and in section 6, I have used a potentially better permutation test (Winkler et al. 2014) that I have used and validated for cross-linguistic analyses in subsequent papers (Koplenig 2021; Koplenig, Wolfer & Meyer 2023). Taken together, the analyses show that when we compare a high L2 language with a low L2 language, both are indistinguishable regarding their (morphological or information-theoretic) complexity provided that both languages have a comparable speaker population size.

Now, critiques like KEW could argue that the categorisation of languages into high L2/vehicular and low L2/non-vehicular by the Ethnologue is incorrect. However, then this begs the question: if we do not trust Ethnologue regarding this categorization, why should we trust them regarding the – arguably more challenging – assessment of both the number of L1 and L2 speakers? As the saying goes - you can't have your cake and eat it.

All in all, I do not have the impression that KEW's comment weakens in any way the argumentation laid out in my original paper. Nonetheless, I agree that a further study using both better data and better methods would certainly be desirable, since testing for a link between language and social structure is turning out to be more complex than I once thought, as recently summarised in an excellent review on the subject (Bromham 2022). As it turns out, there is: in a recent paper, Shcherbakova et al. (2023) use an exciting new database of grammatical features of unprecedented size (Skirgård et al. 2023) in conjunction with novel state-of-the-art statistical methods to account for the effects of genealogical and geographic non-independence of languages (Dinnage, Skeels & Cardillo 2020). While Shcherbakova et al. (2023) find – pretty much in line with the results presented in my paper – a weak but positive effect of speaker population size on complexity, they do not find any negative effect of the L2 proportion on language complexity. The title of their paper neatly sums this up nicely: "Societies of strangers do not speak grammatically simpler languages" (p. 1).

**Acknowledgements**

I very much thank Vera Kempe, the thorough reviewer of my original paper. I also thank Sascha Wolfer and Peter Meyer for input and feedback and Henri Kauhanen, Sarah Einhaus and George Walkden for their kind and honest response to my notification of my intention to submit this rejoinder.

**Funding**

I received no external funding.

**Data availability**

The data provided by Kauhanen et al. (2023) were used for all analyses except for the repeated random draws (Section 2), for which I used the Ethnologue Global Dataset (20th edition), which is a licensed product with restricted terms of use (Personal Research License).

**Code availability**

Commented Stata 17 code to reproduce all results and log files containing all output are available at https://osf.io/fypx5/.